\documentclass[lettersize,journal]{IEEEtran}
\usepackage{amsmath,amsfonts}
\usepackage{algorithmic}
\usepackage{algorithm}
\usepackage{array}
\usepackage[caption=false,font=normalsize,labelfont=sf,textfont=sf]{subfig}
\usepackage{textcomp}
\usepackage{stfloats}
\usepackage{url}
\usepackage{verbatim}
\usepackage{graphicx}
\usepackage{cite}
\usepackage{microtype}
\usepackage{multirow} 
\usepackage{booktabs} 
\usepackage{xcolor}
\usepackage{color}

\usepackage{dsfont}
\definecolor{dg}{rgb}{0,0.694,0.298}
\definecolor{purple}{rgb}{0.4,0.176,0.569}
\definecolor{royalblue}{RGB}{65,105,225}
\usepackage{pifont}
\makeatletter
\DeclareRobustCommand\onedot{\futurelet\@let@token\@onedot}
\def\@onedot{\ifx\@let@token.\else.\null\fi\xspace}

\makeatother

\definecolor{americanrose}{rgb}{1.0, 0.01, 0.24}

\usepackage[capitalize]{cleveref}
\crefname{section}{Sec.}{Secs.}
\Crefname{section}{Section}{Sections}
\Crefname{table}{Table}{Tables}
\crefname{table}{Tab.}{Tabs.}

\begin{document}

\title{Denoising-Contrastive  Alignment for Continuous Sign Language Recognition}

\author{
        Leming Guo,
        Wanli Xue\textsuperscript{*},~\IEEEmembership{Member,~IEEE,}
        and Shengyong Chen,~\IEEEmembership{Senior Member,~IEEE,}
        
\thanks{This work was supported in part by the National Natural Science Foundation of China under Grant 62376197, Grant 62020106004 and Grant 92048301; in part by the Tianjin Science and Technology Program under Grant 23JCYBJC00360.}
\thanks{L. Guo, W. Xue, and S. Chen are with the School of Computer Science and Engineering, Tianjin University of Technology, Tianjin 300384, China.}
\thanks{\textsuperscript{*}W. Xue is the corresponding author (xuewanli@email.tjut.edu.cn).}
}

\markboth{Submitted to IEEE TRANSACTIONS ON IMAGE PROCESSING. Do not distribute}%
{Denoising-Contrastive Alignment for Continuous Sign Language Recognition}

\maketitle
\begin{abstract}
Continuous sign language recognition (CSLR) aims to recognize signs in untrimmed sign language videos to textual glosses.
A key challenge of CSLR is achieving effective cross-modality alignment between video and gloss sequences to enhance video representation.
However, current cross-modality alignment paradigms often neglect the role of textual grammar to guide the video representation in learning global temporal context, which adversely affects recognition performance.
To tackle this limitation, we propose a \textbf{D}enoising-\textbf{C}ontrastive \textbf{A}lignment (DCA) paradigm.
DCA creatively leverages textual grammar to enhance video representations through two complementary approaches: modeling the instance correspondence between signs and glosses from a discrimination perspective and aligning their global context from a generative perspective.
%
%
Specifically, DCA accomplishes flexible instance-level correspondence between signs and glosses using a contrastive loss.
Building on this, DCA models global context alignment between the video and gloss sequences by denoising the gloss representation from noise, guided by video representation.
%
Additionally, DCA introduces gradient modulation to optimize the alignment and recognition gradients, ensuring a more effective learning process. 
%
By integrating gloss-wise and global context knowledge, DCA significantly enhances video representations for CSLR tasks.
%
%
%
Experimental results across public benchmarks validate the effectiveness of DCA and confirm its video representation enhancement feasibility. 
\end{abstract}
    
\begin{IEEEkeywords}
Continuous Sign Language Recognition, Denoising-Diffusion, Global Context Alignment, Instance Corresponding.
\end{IEEEkeywords}
\section{Introduction}
\label{sec:intro}

Sign language is a convenient means of communication between hearing-impaired people, which is a typical visual language.
If hearing people can understand sign language, it will be effective in promoting the active integration of hearing-impaired people into society, which is a difficult challenge.
%
Continuous Sign Language Recognition (CSLR) research aims to address the above challenge. As a representative study of AI serving social good, CSLR refers to recognizing sign language in untrimmed video to textual glosses\footnote{Gloss: each word from the text sentence corresponding to the sign language video.}, which facilitates hearing people understanding sign language.


\begin{figure}[!htbp]
	\centering
    \includegraphics[width=1.0\linewidth]{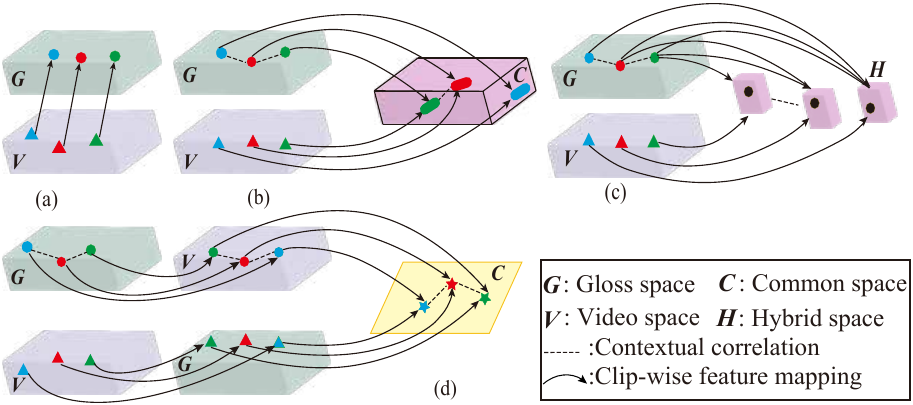}
	
 \caption{Cross-modality alignment paradigms investigated in CSLR. (a) Video clip$\to$individual gloss mapping in the gloss space. (b) Two modalities' distribution close in the high-dimensional common latent space. (c)  Video clip and glosses of previous time steps mapping in the multi-hybrid spaces. (d)
 Two modalities' clip-wise features mutual mapping and are projected into a low-dimensional common space.
 }
 \label{fig:fig1}
\end{figure}
Currently, CSLR is modeled as the typical cross-modality recognition task. Therefore, effective cross-modalities alignment paradigms are needed to enhance the video representation~\cite{collamultilingual2023,Xue2023AlleviatingDI,CVTSLR2023,guo2024gloss}, which can significantly affect the recognition results.
These alignment paradigms mainly fall into the following categories: \ding{182} Employing Connectionist Temporal Classification (CTC) to map video representations to gloss space~\cite{Radial_2022,hu2023continuous,gan2024signgraph}~(Figure~\ref{fig:fig1}(a)). 
\ding{183} 
Employ the language model to learn gloss context, and extract gloss representations. Then map both video and gloss representations~ to a common high-dimensional space to close their distributions~\cite{pu2020boosting,CVTSLR2023, Guo_2023_CVPR, guo2024gloss}~(Figure~\ref{fig:fig1}(b)). 
\ding{184} Mapping video representations at each time step with the gloss representation~(extracted from pre-trained language model) of all previous time steps into multiple high-dimensional hybrid spaces~\cite{Zhang_2023_ICCV}~(Figure~\ref{fig:fig1}(c)).

Specifically, for the paradigm~\ding{182}, video representations be mapped to close the one-hot representation of the corresponding gloss, the video encoder can not learn the gloss semantic to refine itself. 
Although both the video encoder of paradigm~\ding{183} and~\ding{184} have gloss semantics prior, they exclusively learn adjacent gloss context, neglecting the global context alignment to learn the textual grammar.
Thus their video encoder fails to receive the global gloss context to refine themselves.
%
%

Inspired by natural language processing~\cite{clark-etal-2019-bert}, we attribute the textual grammar guidance need to satisfy three key points:~\textit{(1) modeling the language model to capture the textual grammar of the gloss sequence, which knows each gloss semantic and the global context of the gloss sequence.}~\textit{(2) modeling the instance correspondence between signs and glosses to guide the video encoder in signs semantics learning.}
~\textit{(3) modeling the global context alignment between video sequence and gloss sequence to guide the video encoder in global temporal context learning.} 

%
To effectively learn the textual grammar, we propose a novel \textbf{D}enoising-\textbf{C}ontrastive \textbf{A}lignment~(DCA) paradigm.
This paradigm models the sign-gloss instance correspondence from a contrastive perspective, while also aligning the global context between video and gloss sequences from a generative perspective.
%
Specifically, DCA first uses language model fine-tuning to capture the semantic meaning of glosses and the global context of the gloss sequence.
Then, DCA performs a Contrastive Instance Alignment to encourage each sign to match with its most semantic relevant gloss, pulling them closer.
This process allows the video encoder to learn the semantics of each sign.
Building on this, DCA proposes the Denoising-Diffusion Alignment, which encourages video sequence representations to generate gloss sequence representations from noise in low-dimensional latent space. 
Generative capacity is considered one of the highest manifestations of learning~\cite{johnson2018image,hudson2024soda}.
During the generative denoising process, this alignment enables the video encoder to implicitly align video sequence representations with gloss sequence representations. 
Furthermore, this denoising aids the video encoder in developing a deep understanding of essential factors necessary for generating accurate gloss sequence representations. 
These factors include the order and global context knowledge of glosses, which guide the video encoder in reconstructing the global temporal context of all video clips, as shown in Figure~\ref{fig:fig1}(d).
Finally, we introduce a gradient modulation to optimize the angle between the alignment gradient and the recognition gradient to avoid optimization conflict.

%
%
In summary, this study focuses on textual grammar learning to enhance video representations by addressing two key aspects: instance correspondence learning between signs and glosses, and global context alignment between video and gloss sequences.
Experimental results on public CSLR benchmarks demonstrate the effectiveness of the proposed DCA approach, achieving state-of-the-art performance and excelling in global temporal context learning.

The main contributions are summarized as follows:
\begin{itemize}

\item Orthogonal to the cross-modality alignment paradigms in
CSLR, we propose a novel {D}enoising-{C}ontrastive {A}lignment~(DCA) approach.
DCA integrates textual grammar guidance and decomposes it into: instance correspondence learning and global context alignment, which are proposed for the first time in CSLR.
%
%
\item We introduce the Contrastive Instance Alignment to model the sign-gloss instance correspondence in a discrimination manner. Furthermore, we propose the Denoising-Diffusion Alignment to align the global context between video and gloss sequences by generating gloss representations from noise based on video representations.
Together, these techniques enable mutual learning, enhancing both local and global alignment.

\item Experimental results demonstrate that DCA achieves state-of-the-art recognition performance, enhances the generalization of video representations, and achieves satisfactory global temporal context learning.
\end{itemize}

\section{Related Work}

\subsection{Continuous Sign Language Recognition}
\label{sec_realred_cslr}
The CSLR aims to recognize signs in a video corresponding to several glosses, where the order of glosses is consistent with the signs.
Due to weak sentence-level annotation~(lacking segmentation ground-truth for each sign) and small-scale data being available in current CSLR benchmarks, many state-of-the-art methods~\cite{Guo_2023_CVPR,Zhang_2023_ICCV,gan2024signgraph} exploit
the connectionist temporal classication (CTC) \cite{graves2006connectionist} to accurately map each video clip to the corresponding gloss by maximizing the probabilities of all alignment paths between video clips and glosses. 
Based on it, some methods exploit pre-captured pose heatmaps~\cite{zuo2022c2slr} or body keypoints~\cite{chen2022two}, model movements trajectories~\cite{Hu2022SelfEmphasizingNF,hu2023continuous} or build graphs~\cite{gan2024signgraph} among frames,
to further enhance video representations.
However, these methods ignore the problem of the CTC conditional independence assumption, which only achieves video clip$\to$individual gloss mapping, and lacks gloss context learning. 
To solve this problem, ~\cite{chen2022simple,CVTSLR2023,guo2024gloss} further exploit the language model or VAE model or provide extensive gloss context supervision, subsequently map the visual representation to the high-dimensional gloss space, and employ constraints to close their distributions.
Meanwhile, C$^{2}$ST \cite{Zhang_2023_ICCV} recurrently fuses gloss representations from all previous time steps with the current time visual representation, which conducts multiple hybrid spaces to inject the context of two modalities.
However, these methods only focus on guiding video clip representations to learn gloss context, ignoring guiding video clips by the textual grammar to learn their global temporal context alignment.
%

In this study, DCA guides the video encoder in learning sign semantics via the flexible correspondence between video clips.
Besides the video encoder also can be guided to learn the global context of gloss sequence via denoising the noisy gloss sequence representation based on the video.

\subsection{Corss-modality alignment}
In CSLR, due to the weak annotation of benchmarks, almost all methods adopt the CTC loss function to map the visual representations to textual gloss space for cross-modality alignment~\cite{pu2020boosting,min2021visual,chen2022two,CVTSLR2023,guo2024gloss}.
Based on the CTC, some methods also employ the cross-attention operation~\cite{Guo_2023_CVPR}, dynamic time warping (DTW)~\cite{pu2020boosting}, and contrastive learning method~\cite{CVTSLR2023} to provide gloss context supervision.
Current advanced cross-modal alignment are contrastive learning methods~\cite{li-etal-2021-unimo,clip2021} as well as diffusion models~\cite{Huang2023DiffDisEG,Jin2023DiffusionRetGT}.
Specifically, CLIP~\cite{clip2021} conducts a contrastive learning paradigm to align each word and image.
DiffDis~\cite{Huang2023DiffDisEG} employs the powerful diffusion model to perform superior image-text alignment. 
Inspired by the above, we adopt contrastive learning to achieve instance-level cross-modal alignment and denoising-diffusion paradigms to achieve global cross-modal alignment. 

\subsection{Denoising Diffusion Model}
\label{sec_realred_diffusion_model}
The denoising diffusion model~(DDM) incorporates a forward Gaussian diffusion noising process and a reverse denoising generation process, which can refine the generated objects iteratively starting from Gaussian noise.
Recent works~\cite{Jin2023DiffusionRetGT,Huang2023DiffDisEG} have presented that the latent space of the DDM can present the cross-modalities alignment.
Specifically, DiffusionRet~\cite{Jin2023DiffusionRetGT} models text-video retrieval as a process of gradually generating joint distribution from noise.
l-DAE~\cite{Chen2024DeconstructingDD} transforms the DDM to a denoising autoencoder to perform self-supervised learning for recognition-oriented representations and concludes that the representation capability of DDM is mainly gained by the denoising process, rather than a diffusion process.
Besides, DIVA~\cite{Wang2024DiffusionFH} employs the generative feedback from text-to-image diffusion models to refine the fine-grained visual abilities of CLIP.
In this work, we formulate the global alignment between video and gloss sequences as a generative denoising process and use the alignment to supervise video representation.

\section{Method}
\label{sec_method}

\begin{figure*}
  \centering
    \includegraphics[width=0.9\linewidth]{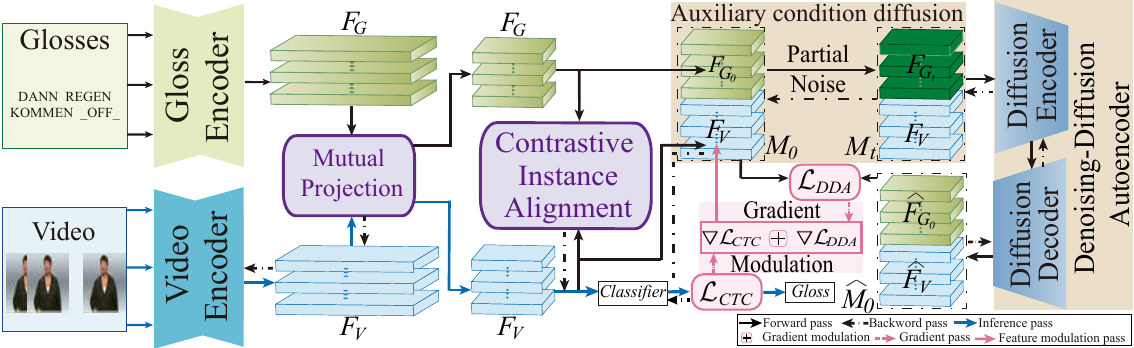}

  \caption{Illustration of the proposed Denoising-Contrastive Alignment~(DCA). 
  We begin by using the Contrastive instance Alignment to model the sign-gloss instance correspondence to learn sign semantics.
  This approach encourages each sign to match with its most semantically relevant gloss, 
  Next, we propose Denoising-Diffusion Alignment to align the global context between video and gloss sequences.
  This technique guides video sequence representations to reconstruct gloss 
  sequence representations from noise.
  The learned instance and global gloss semantics then supervise the video encoder, helping refine the global temporal context of video representations.
  Finally, we introduce gradient modulation to adjust the optimization angle between the alignment gradient and the recognition gradient to avoid optimization conflict.
  }
  \label{fig:DCA}
\end{figure*}

In this section, we first introduce the general CSLR framework.
Then, we propose the Contrastive instance alignment~(CIA), which models the flexible sign-gloss instance correspondence to learn the sign semantic in Sec.\ref{sec:CIA}. After that, we propose the Denoising-Diffusion Alignment~(DDA), which aligns the global context between video and gloss sequences in a generative denoising manner in Sec.\ref{sec:DDA}.
Moreover, we conduct an objective function to optimize the gradient optimization direction among CIA, DDA, and CTC loss functions to modulate the alignment and recognition in Sec.\ref{sec:Optimization}.
Finally, we analyze the correlation between the Denoising-Diffusion Alignment and the CSLR task~in Sec.\ref{sec:revisit}.
The illustration of the proposed Denoising-
Contrastive Alignmen~(DCA) is presented in Figure~\ref{fig:DCA}. 

\subsection{General CSLR Framework}
\label{sec:clsr_baseline}

Continuous sign language recognition (CSLR) framework~\cite{CVTSLR2023,Zhang_2023_ICCV} embraces the paradigm that comprises a video encoder, a classification module, and a cross-modality alignment function.

\noindent\textbf{Video encoder.} The video encoder~$\Phi_{VEnc}$ contains a spatial perception module and a temporal perception module.
Formally, given a $T$ frames sign language video $\mathcal{X} =\{x_i\}_{i=1}^T$, 
the spatial perception module first extracts spatial features $F_{sp}=\{f_{sp}^i\}_{i=1}^T$ from $\mathcal{X}$.
Subsequently, the temporal perception module learns sign-specific knowledge and contextual correlation to extract video representations $F_{V}=\{f_{V}^i\}_{i=1}^{T'} \in {\mathbb{R}^{({T'}) \times d}}$, which is the video encoder's output.
Moreover, $F_{V}$ will be fed into the classifier~$\Phi_{Z}$ to predict corresponding logits ${Z_{V}}=\{z_{V}^i\}_{i=1}^{T'}$.
Finally, the cross-modality alignment function learns the mapping $p\left({g_i|clip ;\theta} \right)$. 
\emph{clip}=$\{x_i\}_{i=1}^{T'}, {T'}<T$ is the video clip in a video, and $\theta$ indicates video encoder's parameters.

\noindent \textbf{Connectionist Temporal Classification}.

For sign language video $\mathcal{X} =\{x_i\}_{i=1}^T$ and its corresponding gloss sequence $\mathcal{G} =\{g_i\}_{i=1}^L$, the Connectionist Temporal Classification~\cite{graves2006connectionist} is employed as~$\mathcal{H}$, which maps unsegmented video clips $\{x_i\}_{i=1}^{T'}$ to glosses sequence $\{g^i\}_{i=1}^L$ by summing the probabilities of all feasible alignment paths~$\pi$:
\begin{equation}
    {\mathcal{H}} =  - \log p\left( {\mathcal{G}|{x_i};\theta} \right) =  - \log \left( {\sum\limits_\pi  {p\left( {\pi |{x_i};\theta} \right)} } \right),
\end{equation}

\noindent where ${p\left( {\pi |{x_i};\theta} \right)}$ is calculated by CTC: $p\left( {\pi |{x_i};\theta} \right) = \prod\limits_{i} {p\left( {{\pi _i}|{x_i};\theta} \right)}$, the probabilities $P_\theta=softmax(Z_V)$ can be calculated via a \emph{softmax} function to the video encoder's  logits $Z_V$.
The CTC has been validated to achieve superior video clip$\to$individual gloss alignment.

\noindent \textbf{Baseline method.} It is worth noticing that the video encoder equipped the classifier and the CTC alignment function~$\mathcal{H}$ standing for the baseline in this work, which is the same as the other CSLR methods~\cite{hu2023continuous,Guo_2023_CVPR,Zhang_2023_ICCV}.

\subsection{Textual grammar capturing}
\label{sec:Textualgrammar}
Inspired by the natural language processing~\cite{clark-etal-2019-bert,marcheggiani,kenton2019bert}, we divide textual grammar learning into two components: each gloss semantic learning and global context learning of gloss sequence.
To achieve this, we first fine-tune the mBART model\footnote{https://huggingface.co/facebook/mbart-large-cc25}, using sign language data~\cite{chen2022two, CVTSLR2023} to extract the semantics of each gloss.
This mBART, pre-trained on the CC25 corpus\footnote{https://commoncrawl.org/} has enriched knowledge of sentence-level global context and the discriminative semantics of individual words.
After fine-tuning, mBART can transfer the semantics of words to their corresponding glosses~((where glosses are a subset of natural language words)), while reserving global context learning prior. 
To further learn the global context of gloss sequence, we adopt a masked language model (MLM) scheme, similar to GPGN~\cite{guo2024gloss}, where approximately 50\% of the glosses in a sentence are masked to predict the remaining ones. 
The optimized mBART model, after fine-tuning, is used as our Gloss Encoder.
Given the sign language gloss sequence~$\mathcal{G} =\{g^i\}_{i=1}^L$, the mBART will extract the
gloss sequence representation~$F_{G}\in {\mathbb{R}^{L \times d}}$.

\subsection{Contrastive instance alignment}
\label{sec:CIA}
\noindent\textbf{Video-Gloss mutual projection.}
The current CSLR methods achieve video clip$\to$gloss mapping in the high-dimensional space~(illustrated in Figure~\ref{fig:fig1}(b) and (c)).
However, this mapping manner faces a large modality bias, making it difficult to transfer information effectively~\cite{Chen2022MSDNMS}.
To address this issue, we propose a video-gloss mutual projection to narrow the bias and 
facilitate smoother context transfer between the two modalities.
Given the pair of visual sequence representation $F_{V}$ and gloss sequence representation $F_{G}$, which cannot fully capture each other's semantics, we first project each representation into the feature space of the other modality using V-L mapper~\cite{chen2022simple}.
Next, both representations are projected into a low-dimensional latent space through two linear layers~(illustrated in Figure~\ref{fig:fig1}(d)).

\begin{equation}
    \begin{aligned}
        &F_{G}=VL(F_{G}), F_{V}=VL(F_{V}),
    \\
    &F_{G}=Linear(F_{G}), F_{V}=Linear(F_{V}).
    \end{aligned}
\end{equation}

After the above projection, the bias between video and gloss modalities can be reduced, and the semantics of the two modalities can be abstracted in low-dimensional space, which can more easily exchange contextual and fine-grained knowledge, improving their consistency.

\noindent \textbf{Contrastive instance alignment.}
Enforcing to learn the sign semantic, based on the proposed mutual projection, we conduct a Contrastive Instance Alignment (CIA) to model the flexible sign-gloss instance correspondence.
Mathematically, given a pair of mini-batch video representations and corresponding glosses representations, we can naturally conduct pairs of video clips and glosses~$\left\{ {\left( {{F_V},{F_G}} \right)} \right\}$. 
Then the CIA first computes the similarity matrix of each video clip and gloss, ${E^{(i,j)}} = {F_V} \cdot F_G^{\rm T} \in {R^{T' \times L}}$. 
Then compute the Softmax scores of each row to re-weight similarities to obtain the correspondence $S^{(i,j)}$ of the instance approximate between signs and glosses.
Finally, the contrastive instance alignment is formulated as:

\begin{equation}
    {\mathcal{L}_{CIA}} =  - \frac{1}{{2B}}\sum\limits_{i = 1}^B {\log \frac{{\exp ({S^{(i,j)}}/\tau )}}{{\sum\nolimits_{j = 1}^B {\exp ({S^{(i,j)}}/\tau )} }}},
\end{equation}

\noindent where $\tau=0.07$ is a smooth temperature coefficient, $B$ denotes the pairs of video clips and glosses. The ${\mathcal{L}_{CIA}}$ function will be optimized to minimize its value, thereby enlarging the similarity between signs and glosses.
As a result, the video encoder will be encouraged to find the most corresponding gloss representation for its video clips' representations and pull them closer to achieve the sign$\to$gloss instance alignment.
This design enhances the video encoder's representation by allowing the gloss sequence's sign semantics to be reflected.

\subsection{Denoising-Diffusion Alignment}
\label{sec:DDA}


\noindent\textbf{Auxiliary condition diffusion.}
\label{sec:acdn}
After the Video-Gloss mutual projection, $F_{G}$ will be
used as prompt, and be concatenated to $F_{V}$ to generate a video-gloss bimodal representation~$M=[{F_{G}};{F_{V}}]=[f_{G}^1,...,f_{G}^{T'},f_{V}^1,...,f_{V}^L]\in {\mathbb{R}^{(T' + L) \times d}}$.
Where $T'$ and $L$ denote the sequence length of visual representation and gloss sequence representation, respectively.
Therefore, $M$ can be regarded as incorporating the gloss sequence's global context into the approximate posterior $q_\phi(M|\mathcal{X}, \mathcal{G})$, where $\mathcal{X}$, $\mathcal{G}$ denotes video and gloss sequences.
Next, we adopt the DDM~\cite{DDPM_2020,2021ddim} forward process to incrementally add multi-level Gaussian noise to the gloss part in bimodal representation $M_0=M$ with a Markov
chain manner~${F_{G}}_0,...,{F_{G}}_T$, ${F_{G}}_0={F_{G}}$. The step from ${F_{G}}_{t-1}$ to ${F_{G}}_t$) is shown as follows: 
\begin{equation}
    q\left({{{F_{G}}_t}|{{F_{G}}_{t-1}}} \right) = \mathcal{N}\left({{F_{G}}_t};\sqrt{1 - {\beta _t}}{{F_{G}}_{t - 1}}, {\beta _t}\mathrm{I}\right).
\label{Eq:xt_xt_1}
\end{equation}
Thus, given ${F_{G}}_0$, ${F_{G}}_t$ can be empirically redefined as: 
\begin{equation}
\begin{aligned}
    {F_{G}}_t = \sqrt{{{\overline \alpha  }_t}}{{F_{G}}_{0}} + \sqrt{1-{{\overline \alpha  }_t}}\epsilon,
\end{aligned}
\label{Eq:Mt_M0}
\end{equation}

\noindent where $\varepsilon$ stands for Gaussian noises, and we empirically set diffusion steps including~$T$, ${{\beta _t} \in (0,1)} _{t = 1}^t$, ${\alpha _t}: = 1 - {\beta _t}$, ${\overline \alpha  _t}: = \prod\nolimits_s {{\alpha _s}}$. 
Additionally, the inverted noise schedule~\cite{Hudson_2024_CVPR} is adopted. It focuses on the use of moderate noise levels, which helps recognition representation learning.
In particular, ${M_t}=[{F_{G}}_t;{F_{V}}]$ denotes the noisy gloss sequence representation connected with the video representation.

\noindent \textbf{Denoising-diffusion autoencoder.}
\label{sec:ddae}
To model the global context alignment between video and gloss sequences, we conduct a denoising-diffusion autoencoder~\cite{Chen2024DeconstructingDD} to generate the gloss sequence representation from noise, based on the video sequence representation.
%
The diffusion-denoising autoencoder contains a diffusion encoder~${\Phi_{DEnc} }$ and a diffusion decoder~${\Phi_{DDec}}$.
Inspired by the framework of l-DAE~\cite{Chen2024DeconstructingDD}, both the encoder and decoder have the same architecture, and the decoder is deeper than the encoder.
Given the gloss-part noised bimodal representation $M_t$,
the diffusion encoder~$\Phi_{DEnc}(M_t,t)$ embeds the noised gloss sequence representation and learns the context of the video sequence representation from the video sequence representation, outputting the latent features~$F_H$.
${\Phi_{DEnc}}$ is optimized to denoise the noise part of $M_t$ based on the video sequence representation, preserving the most relevant information between the initial bimodal representation~$M_0$ and the denoising result.
The diffusion decoder~$\Phi_{DDec}(M_t,t)$ then decodes~$F_H$ to the denoising output~$\hat{M}_0={\Phi_{DDec} }\left( {{M_t},t} \right)$. 
This model is optimized to make the tractable variational lower-bound~$\mathcal{L}_{VLB} \le \mathbb{E}[{-logp_\theta(M_0)}]$, ensuring effective denoising of $M_t$ and an approximate of the clean initial bimodal representation~$M_0$~\cite{DDPM_2020,2021ddim}.
$\mathcal{L}_{VLB}$ is the negative Evidence Lower Bound:$-KL\left( {{q_\theta }(M_0|\mathcal{X},\mathcal{G})||{p_\theta }(M_0|\mathcal{X})} \right) + {E_{{q_\theta }(M_0|\mathcal{X},\mathcal{G})}}\left[ {\log {p_\theta }(\mathcal{G}|M_0,\mathcal{X})} \right] \le \log {p_\theta }(\mathcal{G}|\mathcal{X})$, where $G$, and $X$ denote the gloss and video sequences. 
We follow~\cite{DDPM_2020}
to simplify~$\mathcal{L}_{VLB} \le \mathbb{E}[{-logp_\theta(M_0)}]$ to the mean squared error.
The simplified objective is defined as $\mathcal{L}_{DDA}$:
\begin{equation}
    \begin{aligned}
    &\mathcal{L}_{DDA}= {\mathbb{E}_{{M_0} \sim q({M_0}),{{{\hat M}_{0}} ,t \sim [1,T]}}{\left\| {{M_{0}}  - \hat{M}_{0}} \right\|^2}}.
    \end{aligned}
\label{eq:L_diff}
\end{equation}


When optimizing Eq.~\ref{eq:L_diff}, the noise is denoised based on the video sequence representation to generate the gloss sequence representation. 
The denoising-diffusion autoencoder implicitly aligns the video sequence representation with the gloss, facilitating the learning of joint video-text feature space, similar to other video-gloss alignment methods.
In this joint feature space, both the autoencoder and the video encoder must infer the gloss sequence's global context.
Through these approaches, the autoencoder aids the video encoder in having a deep understanding of the gloss order and the global context that defines the gloss sequence representation.
As a result, the video encoder is implicitly optimized to refine its global temporal context, aligning it with the gloss global context.


\subsection{Correlation between Denoising-diffusion alignment and CSLR.}
\label{sec:revisit}
Formally, given a $T$ frames sign language video $\mathcal{X} =\{x_i\}_{i=1}^T$, and its corresponding gloss sequence with $L$ glosses $\mathcal{G} =\{g_i\}_{i=1}^L$, we introduce the video representation $F_{V}$~(see Sec.3.1) as latent variables $z$ to help model the conditional probability of CSLR:
\begin{equation}
    p\left( {\mathcal{G}|\mathcal{X}} \right) = \int\limits_z {p\left( {\mathcal{G},z|\mathcal{X}} \right)} dz = \int\limits_z {p\left( {\mathcal{G}|z,\mathcal{X}} \right)} p\left( {z|\mathcal{X}} \right)dz.
    \label{eq:revisit}
\end{equation}
In Eq.\ref{eq:revisit}, we find that latent variable $z$ is key to predicting glosses, and it serves as a springboard to help narrow the bias between video and gloss modalities. 
Furthermore, we theoretically try to optimize the true posterior probability~$p_\theta(z|\mathcal{X},\mathcal{G})$ to get the ideal $z$.
Due to the unsolvable property of the true posterior distribution $p_\theta(z|\mathcal{X},\mathcal{G})$, a variational approximate posterior distribution $q_\theta(z|\mathcal{X},\mathcal{G})$ is conducted to train the CSLR model, which is optimized by the Evidence Lower Bound:
$-KL\left( {{q_\theta }(z|\mathcal{X},\mathcal{G})||{p_\theta }(z|\mathcal{X})} \right) + {E_{{q_\theta }(z|\mathcal{X},\mathcal{G})}}\left[ {\log {p_\theta }(\mathcal{G}|z,\mathcal{X})} \right] \le \log {p_\theta }(\mathcal{G}|\mathcal{X})$. 
Notice that the Evidence Lower Bound optimization is the same as the  
the negative tractable variational lower-bound~$\mathcal{L}_{VLB} \le \mathbb{E}[{-logp_\theta(M_0)}]$ in Sec.\ref{sec:ddae}, which can be achieved by optimizing the denoising-diffusion autoencoder.
By optimizing $\mathcal{L}_{VLB}$, the prior $p_\theta(z|\mathcal{X})$ can effectively capture bi-modal information from the posterior distribution $q_\phi(z|\mathcal{X},\mathcal{G})$, to enhance the latent variable $z$.
%

\subsection{Optimization}
\label{sec:Optimization}
The objective of our method~$\mathcal{L}_{DCA}$ contains the objective function~$\mathcal{L}_{DDA}$ and $\mathcal{L}_{CIA}$ and $\mathcal{L}_{CTC}$.
Specifically, combining the $\mathcal{L}_{DDA}$ and $\mathcal{L}_{CIA}$ ensures the video encoder has an effective capturing of the sign semantics and the global context within the textual gloss modality.
Besides, $\mathcal{L}_{CTC}$, the connectionist temporal classification, can establish the many-to-one corresponding between multiple video frames and one gloss, achieving the gloss classification.
Above all, our objective $\mathcal{L}_{DCA}$ is defined as follows:
\begin{equation}
    \mathcal{L}_{DCA} = \mathcal{L}_{CTC}+ \mathcal{L}_{CIA} + \gamma_1\mathcal{L}_{DDA},
    \label{Eq:DCA}
\end{equation}
\noindent where $\gamma_1$ is a hyperparameter for balancing the contribution.
 
\noindent \textbf{Gradient Modulation.}
During $\mathcal{L}_{DCA}$ optimization, the video encoder gets the sign-gloss correlation gradients, and the global contextual knowledge gradients, provided from the denoising-diffusion autoencoder feedback, a modulation is needed to optimize the angle with CTC classification gradients.
We consider video-gloss alignments and global contextual knowledge gradients are come from the generative representation $\hat{F}_V$.
Inspired by the domain adaptation task~\cite{GH2024}, we modulate the CTC classification gradient optimization direction of the $F_{V}$ to force it to fit $\hat{F}_V$ by:
\begin{equation}
\begin{aligned}
    {g_o} = {g_o} + {g_a} - \delta ({g_o}^{\rm T}{g_a} < 0)\frac{{{g_o}^{\rm T}{g_a}}}{{{{\left\| {{g_a}} \right\|}^2}}}{g_a} 
    \\ 
    - \delta ({g_a}^{\rm T}{g_o} < 0)\frac{{{g_a}^{\rm T}{g_o}}}{{{{\left\| {{g_o}} \right\|}^2}}}{g_o},
\end{aligned}
\end{equation}
\noindent where ${g_o}$ denotes gradients of $F_V$, and ${g_a}$ indicates gradients of $\hat{F}_V$. If $\delta$ is true, its value is 1, else is 0.


\section{Experiments}
\label{Experiments}

\subsection{Datasets and Evaluation}
\label{sec_Data_implement}
\noindent\textbf{PHOENIX-2014 \cite{koller2015continuous}.} This benchmark was recorded from public sign language interpreters of weather forecasts, and it delivers 6,842 sentences interpreted by 9 signers, composed of 1,295 sign language gloss vocabulary.
Additionally, the PHOENIX-2014 dataset is officially divided into the train set, dev set, and test set with 5,672, 540, and 629 videos.

\noindent\textbf{PHOENIX-2014T\cite{camgoz2018neural}.} This benchmark is widely utilized for both continuous sign language recognition~(CSLR) task and sign language translation~(SLT) task. 
It contains 1,085 sign language gloss vocabulary for the CSLR task, and 
all videos are officially divided into 7,096, 519, and 642 videos for the training, dev, and test sets, respectively.

\noindent\textbf{CSL-Daily\cite{zhou2021improving}.} This benchmark is a large Chinese CSLR dataset for both continuous sign language recognition~(CSLR) task and sign language translation~(SLT) task. 
Specifically, the CSLR task records 2000 sign language vocabulary and it is officially divided into 18,401, 1,077, and 1,176 videos for the training, dev, and test sets.


\noindent \textbf{Evaluation metric.} 
This work adopts the word error rate~(WER) metric for the CSLR evaluation {and employs the compression of information stored in weights (IIW)~\cite{PIB2022} to present the generalization error~(i.e., the accuracy difference on the test and training datasets) of neural networks.}
Specifically, the WER belongs to the edit distance, which measures the minimum number of substitutions~(\#sub), deletions~(\#del), and insertions~(\#ins) operation needed to convert the predicted gloss sequence to the associated reference gloss sequence. The WER calculation method is as follows:
\begin{equation}
WER = \frac{{{\#sub} + {\#del} + {\#ins}}}{L},
\end{equation}
\noindent where ${\#sub},{\#del},{\#ins}$ are the number of substitutions, deletions, and insertions operation, respectively. Therefore, lower WER values imply better recognition performance.
{For the IIW, with the same training dataset, a lower IIW of a neural network means that the network has the greater ability to generalize test data.}

\subsection{Implementation Details}

\noindent \textbf{Network architecture.}
Specifically, in the video encoder, the Resnet50~\cite{inproceedingsResnet} pre-trained by CLIP~\cite{clip2021} is leveraged to be the spatial perception module and the temporal perception module is empirically set to TLP~\cite{hu2022temporal} equipped with a two-layer Bi-LSTM module.
The feature dimensions of the TLP and Bi-LSTM are set to $1024$.
For the denoising-diffusion autoencoder, we adopt a two-layer Transformer of the gloss encoder~(see Sec.\ref{sec:Textualgrammar}) as the diffusion encoder, and a three-layer Transformer of the gloss encoder as the diffusion decoder.
Shallow layers can minimize the loss of its prior during the training process. SimMIM~\cite{Xie_2022_CVPR} and MAE~\cite{MAE2022} suggest that a shallower layer for the decoder is sufficient for decoding.

\noindent \textbf{Parameter setting.}
For auxiliary condition diffusion, following DDIM~\cite{2021ddim} and ~\cite{Chen2024DeconstructingDD}, the diffusion noising timestep $T$ is set to $600$, and we set $\beta_t$ from $\beta_1=0.0001$ to $\beta_T=0.99$. 
The covariance $\Sigma_\theta$ is ﬁxed and deﬁned as $\Sigma_\theta=\beta_t \mathbb{I}$.  
The hyperparameter $\gamma_1$ in Eq.~\ref{Eq:DCA} are set to $5.0$.

\noindent \textbf{Training and inference process.}
We train the DCA with a batch size of $4$, using the Adam optimizer~\cite{kingma2014adam} with an initial learning rate of $1e-5$ for Resnet50, and $1e-4$ for others, a weight decay factor of $1e-4$, and momentum as $0.9$ and $0.99$ for $80$ epochs.
And the learning rate decays~($0.2$) at $31$ and $61$ epochs.
Hyperparameter $\gamma_2$ is set to 10.
All experiments are implemented in PyTorch and on two A100 GPUs.
The denoising-diffusion autoencoder is dropped in inference. 
The inference process of DCA begins by feeding the test video~$\mathcal{X}$ into the video encoder to generate video representations. 
These representations are then input into the classifier to obtain classification probability scores. Finally, CTC beam search decodes the output and generates the predicted gloss sequence. 
The beam width is set to 2.

\setlength{\tabcolsep}{2pt}
\begin{table}[t]
\centering
\fontsize{9}{12}\selectfont
\caption{Comparison with baseline methods on~{PHOENIX-2014}. ``Group1-4'' corresponds to the cross-modality alignment paradigm shown in Figure~\ref{fig:fig1}(a), (b), (c) and (d), respectively.
The entries denoted by ``$*$'' used extra cues (keypoints).}
\begin{tabular}{c|c|c|c|c|c} 
\toprule
\multirow{2}{*}{Groups} & \multirow{2}{*}{Methods}  &\multicolumn{2}{c|} {Dev~(\%)~$\downarrow$} &\multicolumn{2}{c}{Test~(\%)~$\downarrow$} \cr \cline{3-6}
      &   &del/ins    &WER          &del/ins        &WER    \cr      \midrule
    
     \multirow{13}{*}{Group1}   & VAC\cite{min2021visual}              &7.9/2.5 & {21.2}  &8.4/2.6 & 22.3 \cr
       & SMKD\cite{hao2021self}              &6.8/2.5 & {20.8}  &6.3/2.3 & {21.0} \cr 
       & SEN~\cite{Hu2022SelfEmphasizingNF} &5.8/2.6 & {19.5}  &7.3/4.0 & {21.0} \cr 
       & TLP\cite{hu2022temporal}              &6.3/2.8 & {19.7}  &6.1/2.9 & {20.8} \cr 
       & C$^{2}$SLR$^*$\cite{zuo2022c2slr} &- & {20.5}  &- & {20.4} \cr 
      &  SGN~\cite{SGN_2023} &5.1/3.1 & {19.5}  &5.3/2.8 & {20.2} \cr 
     &   RadialCTC~\cite{Radial_2022} &6.5/2.7 &19.4 &6.1/2.6 &20.2 \cr
     &  CoSign~\cite{Jiao_2023_ICCV} &-&19.7 &- & 20.1 \cr 
    &  SignBERT+~\cite{SignBERT+2023} &4.8/3.7&19.9 &4.2/3.8 & 20.0 \cr 
     &   CorrNet\cite{hu2023continuous}              &5.6/2.8 & {18.8}  &5.7/2.3 & {19.4} \cr 
     &   TwoStream-SLR$^*$\cite{chen2022two}              &- & {18.4}  &- & {18.8} \cr
     &TB-Net\cite{liu2024tb} &- &18.9& -&19.6\cr 
     & SignGraph\cite{gan2024signgraph} &-&18.2&-& 19.1 \cr \hline
  \multirow{4}{*}{Group2}   & CMA\cite{pu2020boosting}            &7.3/2.7 & 21.3  &7.3/2.4 & 21.9 \cr 
     &   CVT-SLR~\cite{CVTSLR2023} &6.4/2.6 & {19.8}  &6.1/2.3 & {20.1} \cr
           &{GPGN}\cite{guo2024gloss}              &5.8/3.6 & {19.9}  &6.3/2.8 & {20.4} \cr 
      &  {CTCA}\cite{Guo_2023_CVPR}              &6.2/2.9 & {19.5}  &6.1/2.6 & {20.1} \cr \hline

     \multirow{1}{*}{Group3}    
     &   C$^{2}$ST\cite{Zhang_2023_ICCV}              &4.2/3.0 & {17.5}  &4.2/3.0 & {17.7} \cr \hline

    {Group4}  &   {DCA}(Ours)          &4.5/2.5 & \textbf{17.3}  &4.4/2.8 & {\textbf{17.7}} \cr
        \bottomrule
    \end{tabular}
    \label{Table:2014}
\end{table}

\setlength{\tabcolsep}{8pt}
\begin{table}[ht]
\centering
\fontsize{9}{12}\selectfont
\caption{{Comparison~($\%$) with baseline methods on the PHOENIX-2014T. The entries denoted by ``$*$'' used extra cues (keypoints). }}
\begin{tabular}{c|cc}
\toprule
\multirow{2}{*}{Methods} & \multicolumn{2}{c}{WER}\cr \cline{2-3} 
&Dev$\%$ & Test$\%$ \cr 
\midrule
{V-L Mapper}~\cite{chen2022simple} &21.9 &22.5\cr 
{TLP}~\cite{hu2022temporal}        & {19.4}   & {21.2} \cr 
{SEN}~\cite{Hu2022SelfEmphasizingNF}        & {19.3}   & {20.7} \cr 
{CorrNet}~\cite{hu2023continuous}        & {18.9}   & {20.5} \cr
{C$^2$SLR}$^*$~\cite{zuo2022c2slr}  &20.2 &{20.4} \cr 
{CVT-SLR}~\cite{Zheng_2023_CVPR}        & {19.4}   & {20.3} \cr 
{CTCA}~\cite{Guo_2023_CVPR}        & {19.3}   & {20.3} \cr
CoSign~\cite{Jiao_2023_ICCV} &19.5 & 20.1 \cr
SignBERT+~\cite{SignBERT+2023} &18.8 & 19.9 \cr
{TwoStream-SLR}$^*$~\cite{chen2022two}        & {17.7}   & {19.3} \cr
C$^{2}$ST\cite{Zhang_2023_ICCV}              & {17.3}  & {18.9} \cr \hline
{DCA}~(Ours)                        &\textbf{17.0} & \textbf{18.5} \cr 
\bottomrule
\end{tabular}
\label{Table:2014t}
\end{table}

\subsection{Comparison with state-of-the-art methods}
\label{sec:sota}
We compare our DCA with several baseline approaches on three public benchmarks,~PHOENIX-2014~\cite{koller2015continuous}, PHOENIX-2014T~\cite{camgoz2018neural}, and CSL-Daily~\cite{zhou2021improving}.
As shown in Table~\ref{Table:2014}, Table~\ref{Table:2014t}, Table~\ref{Table:cslDaily}, our proposed DCA, which models only the RGB cue of sign language video and accompanying gloss sequence, achieves state-of-the-art recognition performances across all three public benchmarks. 
This result emphasizes the effect of the textual grammar guidance in enhancing the video encoder performance.
Table~\ref{Table:2014} illustrates the performance of four cross-modality alignment paradigms~(from Group1 to Group4, corresponding to Figure~\ref{fig:fig1}(a) to Figure~\ref{fig:fig1}(d)).
We observe that CTC-based paradigms (Group 1), while performing significant recognition through strategies such as sign movements capturing~\cite{hu2022temporal,hu2023continuous,liu2024tb,gan2024signgraph}, keypoint supervision~\cite{chen2022two,zuo2022c2slr}, or self-supervised constraints~\cite{min2021visual,hao2021self,SGN_2023}, overlook the importance of gloss semantics and global context learning. 
Additionally, moving from Group 2 to Group 3, we observe that modeling gloss semantics and applying cross-modality alignment schemes to guide the video encoder in learning gloss semantics can effectively embrace recognition gain.
However, some of them have worse performance than the methods in Group 1, likely due to limitations in the video encoder's quality.
This highlights the importance of a strong video encoder representation for achieving good recognition results.
%
Specifically, our DCA outperforms the CTC alignment TwoStream-SLR~\cite{chen2022two} by 1.1\% and 1.1\% WERs on the dev and test set of PHOENIX-2014, and by 0.7\% and 0.8\% WERs on the dev and test set of PHOENIX-2014T~(see Table~\ref{Table:2014t}).
These performance gaps between DCA and CTC-based methods (Group 1) underscore the effect of instance correspondence between signs and glosses, and the global context alignment between video and gloss sequences.

Moreover, our DCA also gains a significant improvement over local gloss context alignment paradigms~(Group2 and Group3).
For instance, C$^{2}$ST~\cite{Zhang_2023_ICCV} injects local gloss context into label prediction to establish the video clip-gloss mapping, which then refines the video representation. 
Remarkably, our DCA exceeds the C$^{2}$ST by 0.2\% on the dev set of PHOENIX-2014 and exceeds it by 0.3\% and 0.4\% WERs on the dev and test set of PHOENIX-2014T~(see Table~\ref{Table:2014t}).
Additionally, DCA surpasses C$^{2}$ST by 0.3\% and 0.5\% on both the dev and test sets on the CSL-Daily~(see Table~\ref{Table:cslDaily}).
These performance gaps highlight the effectiveness of global context alignment, which guides the video encoder to refine its global temporal context in accordance with the gloss sequence's global context.

\setlength{\tabcolsep}{4pt}
\begin{table}[t]
\centering
\fontsize{9}{12}\selectfont
\caption{{Comparison~($\%$) with baseline methods on CSL-Daily. The entries denoted by ``$*$'' used extra cues (keypoints). }} 
\begin{tabular}{c|cc|cc} %
\toprule
\multirow{2}{*}{Methods}  &\multicolumn{2}{c|}{Dev$\%$} &\multicolumn{2}{c}{Test$\%$} \cr \cline{2-5}
&del/ins &WER &del/ins &WER  \cr 
\midrule
{BN-TIN+Transf}~\cite{zhou2021improving}  &13.9/3.4 &33.6 &13.5/3.0 &33.1 \cr 
%
{SEN}~\cite{Hu2022SelfEmphasizingNF}  &- &{31.1} &- &{30.7} \cr 
{SGN}~\cite{SGN_2023}      &10.8/3.1  & {30.4}  &10.7/2.9 & {30.3} \cr
{CorrNet}~\cite{hu2023continuous}  &- &{30.6} &- &{30.1} \cr 
{CTCA}~\cite{Guo_2023_CVPR}  &9.2/2.5 &{31.3} &8.1/2.3 &{29.4} \cr
TB-Net~\cite{liu2024tb} &-&28.4 &-&28.2\cr

CoSign~\cite{Jiao_2023_ICCV} &-&28.1 &- & 27.2 \cr
{TwoStream-SLR}$^*$~\cite{chen2022two}  &- &\textbf{25.4} &- &{\textbf{25.3}} \cr 
C$^{2}$ST\cite{Zhang_2023_ICCV}   &9.3/2.7            & {25.9}  &9.0/2.7 & {25.8} \cr \hline
{DCA}~(Ours)    &9.1/2.8 & {{25.6}}  &9.0/2.1 & {\textbf{25.3}} \cr 
\bottomrule
\end{tabular}
\label{Table:cslDaily}
\end{table}



\setlength{\tabcolsep}{5pt}
\begin{table}[!htbp]
\centering
\fontsize{9}{12}\selectfont
\caption{Ablation study on the DCA on the {PHOENIX-2014}~(WER\%).}
\begin{tabular}{c|c c c|c|c} 
\toprule
        Methods & MP &CIA  & DDA & Dev~(\%)~$\downarrow$ & Test~(\%)~$\downarrow$ \\ \midrule 
        \multirow{5}{*}{Baseline} & - & - & - & 19.5 & 20.4 \\ 
       ~ & \checkmark & \checkmark & - & 18.0~($1.5\%\downarrow$) & 18.8~($1.6\%\downarrow$) \\ 
       ~ & \checkmark & - & \checkmark & 18.0~($1.5\%\downarrow$) & 18.6~($1.8\%\downarrow$) \\ 
       ~ & - & \checkmark & \checkmark & {17.5}~($2.0\%\downarrow$) & {18.2}~($2.2\%\downarrow$) \\ 
        ~ & \checkmark & \checkmark & \checkmark & \textbf{17.3}~($2.2\%\downarrow$) & \textbf{17.7}~($2.7\%\downarrow$) \\ 
        \bottomrule
    \end{tabular}
    \label{Table:ablation}
\end{table}

\setlength{\tabcolsep}{4pt}
\begin{table}[t]
\centering
\fontsize{9}{12}\selectfont
\caption{Ablation study on the Denoising-diffusion autoencoder on the {PHOENIX-2014}. ``Auto-Enc'' and ``Auto-Dec'' denote the Encoder and Decoder of the denoising-diffusion autoencoder, respectively. ``FN'' stands for the full noising process.}
\begin{tabular}{c|c|c} 
\toprule
    {Variants}  & {Dev~(\%)~$\downarrow$} &{Test~(\%)~$\downarrow$} \cr \hline
         
        DCA (Ours)   & {17.3} & {17.7} \\ 
        DCA w/o Auto-Dec  & 18.6~(1.3$\%$$\uparrow$)  & 18.8~(1.1$\%$$\uparrow$) \\ 
        DCA w/o Auto-Eec   & 18.2~(0.9$\%$$\uparrow$)  & 18.6~(0.9$\%$$\uparrow$) \\ 
       DCA w/ FN    & 18.1~(0.8$\%$$\uparrow$)  & 18.6~(0.9$\%$$\uparrow$) \\ 
        \bottomrule
    \end{tabular}
    \label{Table:ablation_ae}
\end{table}

\subsection{Ablation Studies}
\label{sec:ablation}

\noindent \textbf{Ablation on the proposed DCA.}  
We illustrate the effects of the proposed $\mathcal{L}_{DCA}$, Table~\ref{Table:ablation} ablates the ablation studies of the Denoising-Contrastive Alignment~(DCA) on the PHOENIX-2014 benchmark.
We can see that the baseline method~(See \ref{sec:clsr_baseline}) obtains WERs of 19.5\% and 20.4\% on both dev and test sets.
Remarkably, the proposed Contrastive instance alignment~(CIA) yields significant improvement, achieving 18.0\% and 18.8\% on both dev and test sets. 
The introduction of CIA enables the baseline model to flexibly map each sign to its corresponding gloss, helping the video encoder learn gloss semantics and refine its representation.  

Further improvements are achieved by adding Denoising-Diffusion Alignment (DDA), which results in a 1.5\% and 1.8\% improvement in WER on the dev and test sets.
These results highlight the superiority of the global context alignment between video and gloss sequences through the generative denoising process.
This approach implicitly aligns the video sequence with the gloss and helps the video encoder develop a deeper understanding of gloss order and global context.
Additionally, ``MP'' refers to the Video-Gloss mutual projection described in Sec.\ref{sec:CIA}.
When ``MP'' is omitted, we follow the paradigm of Figure~\ref{fig:fig1}(b) to project the video and gloss sequences into a high-dimensional common latent space.
As shown in Table~\ref{Table:ablation}, employing ``MP'' yields significant recognition improvements, reducing 0.2\% and 0.5\% WERs on both dev and test sets.
This indicates that projecting both video and gloss features into each other's feature spaces helps narrow modality bias, enabling the semantics of both modalities to be more effectively captured in a low-dimensional space.

\setlength{\tabcolsep}{4pt}
\begin{table}[!htbp]
\centering
\fontsize{9}{12}\selectfont
\caption{Performance comparison of distinct distribution alignment methods on the {PHOENIX-2014}.}
\begin{tabular}{c|c|c|c} 
\toprule
        Methods & Alignment & Dev~(\%)~$\downarrow$ & Test~(\%)~$\downarrow$ \\ \midrule 
       \multirow{6}{*}{Baseline} & MMD~\cite{Gretton2012AKT} & 18.6 & 19.0 \\ 
        ~ & JMMD~\cite{JMMDlong17a} & 18.3 & 18.6 \\ 
        ~ & NCE~\cite{clip2021} & 17.9 & 18.1 \\ 
        ~ & SimMIM~\cite{Xie_2022_CVPR} & 17.9 & 17.9 \\ 
        ~ & MAE~\cite{MAE2022} & 17.5 & 17.8 \\ 
        ~ & DCA~(Ours) & \textbf{17.3} & \textbf{17.7} \\ 
        \bottomrule
    \end{tabular}
    \label{Table:abl_konwledge}
\end{table}

\noindent \textbf{Ablation on the Denoising-diffusion autoencoder.}  
As shown in Table~\ref{Table:ablation_ae}, when DCA removes the diffusion decoder~(``Auto-Dec''), the remaining diffusion encoder is optimized to learn the added noise at each timestep in the DDM forward process.
This results in a recognition performance drop of 1.3\% and 1.1\% on the dev and test sets, respectively (i.e., an increase in WER).
Further, when DCA removes the diffusion encoder~(``Auto-Enc''), and the remaining diffusion decoder is optimized by $\mathcal{L}_{DCA}$ to approximate $M_0$, a slight performance degradation is observed (0.9\% and 0.9\% increase in WER on the dev and test sets).
The two experiments demonstrate the superiority of the diffusion decoder AND validate that the denoising-driven process is crucial for significant representation learning, as demonstrated in l-DAE~~\cite{Chen2024DeconstructingDD}.
Additionally, ``FN'' refers to the full noising process, where noise is added to both the video and gloss sequence parts of the bimodal representation $M$.
Replacing the partial noising process with ``FN'', leads to a recognition performance drop of 0.8\% and 0.9\% on the dev and test sets (increased WER).
This result demonstrates the effectiveness of the partial noising process for natural modality context learning and for promoting global context alignment between video and gloss sequences.

\noindent \textbf{Ablation on distinct distribution
alignment objects.}
We investigate the superiority of the proposed DCA by comparing it to other distribution alignment methods that aim to achieve local or global context alignment between video and gloss sequences. 
As shown in Table~\ref{Table:abl_konwledge}, 
considering the textual grammar guidance, DCA achieves a more significant performance than MMD~\cite{Gretton2012AKT}, JMMD~\cite{JMMDlong17a}, NCE loss~\cite{clip2021}, SimMIM~\cite{Xie_2022_CVPR}, and  MAE~\cite{MAE2022}.
In particular, while the NCE loss achieves distributions closer to the common sentence and separates distributions for different contexts, its performance is limited by 
ignoring the global alignment.
Additionally, although SimMIM and MAE are generative methods similar to DDM, they cannot simultaneously learn multi-level mask ratios, unlike DDM, which uses multi-level Gaussian noise addition.
Furthermore, both SimMIM and MAE do not model sign-gloss instance correlation well.
These experiments demonstrate that formulating the global temporal context alignment within a denoising-diffusion autoencoder is feasible and achieves promising alignment capacity.


\setlength{\tabcolsep}{2pt}
\begin{table}[!htbp]
\centering
\fontsize{9}{12}\selectfont
\caption{Performance comparison of distinct networks for the autoencoder on the {PHOENIX-2014}.}
\begin{tabular}{c|c|c|c} 
\toprule
        Methods & Networks & Dev~(\%)~$\downarrow$ & Test~(\%)~$\downarrow$ \\ \midrule 
        \multirow{4}{*}{Autoencoder} & pre-trained 
        mBART & \textbf{17.3} & \textbf{17.7} \\ 
        ~ & mBART & 17.9 & 18.2 \\ 
        ~ & BLSTM & 17.5 & 17.8 \\ 
        ~ & 1D U-Net & 17.6 & 17.9 \\ 
        \bottomrule
    \end{tabular}
    \label{Table:abl_auto}
\end{table}


\noindent \textbf{Ablation on distinct networks for the denoising-diffusion autoencoder.}
We investigate the impact of different diffusion encoder and decoder networks, all of which can naturally acquire contextual knowledge. 
``pre-trained mBART'' denotes the mBART language model pre-trained on sign language~(See Sec.\ref{sec:Textualgrammar}).
``mBART'' indicates the mBART language model~\cite{chen2022simple,CVTSLR2023}.
Specifically, ``pre-trained mBART'' model has prior knowledge of sign language grammar, whereas the “mBART” model only has natural language prior knowledge.
As shown in~Table~\ref{Table:abl_auto}, all networks achieve superior performance, indicating that the ability to learn the global semantic correlation between video and gloss sequences contributes to the effectiveness of DDA. 
Notably, the pre-trained mBART model, which retains the prior knowledge of gloss sequence context, outperforms the others, highlighting that more robust contextual learning prior enhances the performance of DDA.



\setlength{\tabcolsep}{5pt}
\begin{table}[!htbp]

\centering
\fontsize{9}{12}\selectfont
\caption{{Ablation study on timestep $T$ factor on the PHOENIX-2014.}}
\begin{tabular}{c|c|c|c|c|c|c} 
\toprule
        timestep $T$ & 1000 & 800 & 600 & 200 & 100 & 50 \\ \midrule 
        Dev~(\%)~$\downarrow$ & 18.1 & 17.5 & \textbf{17.3} & 17.9 & 18.6 & 18.8  \\ \midrule 
        Test~(\%)~$\downarrow$ & 18.3 & 17.9 & \textbf{17.7}  & 18.2 & 18.8 & 19.4\\  
    \bottomrule
    \end{tabular}
    \label{Table:timestep_T}
\end{table}

\begin{figure}[!htbp]
  \centering
    \includegraphics[width=1.0\linewidth]{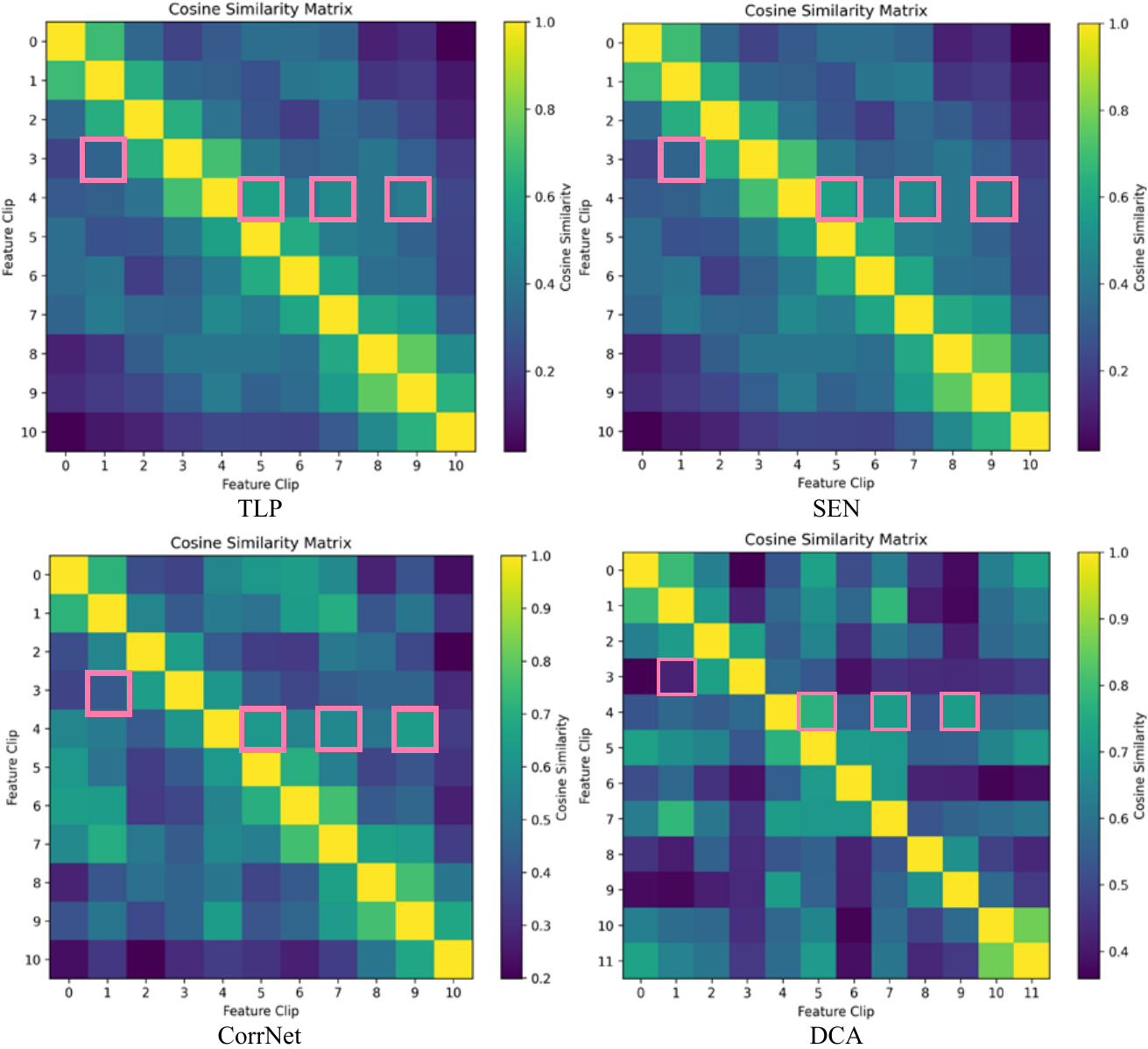}

  \caption{Evaluation for the global temporal context learning over DCA and other SOTA CSLR methods on the PHOENIX-2014 test set.
  }
  \label{fig:DCAcontext}
\end{figure}
\noindent {\textbf{Ablation on the diffusion noising timestep $T$.}
In this section, we gradually decrease the diffusion step for the noising process~(from 1000 to 50).
When the timestep $T$ is small, less noise is added to the gloss sequence representation, and vice versa.
With fewer noise steps, the denoising process becomes easier, but the diffusion decoder learns less gloss context knowledge.
Table~\ref{Table:timestep_T} delivers that the optimal timestep is $600$.
Notably, even when timestep $T$ is $50$ the model still achieves strong performance.
This suggests that when only a small amount of noise is removed during denoising, video representations can still effectively align with the gloss sequence. The limited gloss context knowledge feedback also benefits the video encoder. Furthermore, it appears that applying noise to low-level latent representations does not require a large amount of noise.
On the contrary, when timestep $T$ is too large, the diffusion decoder struggles to infer the gloss sequence representation conditioned on the video.


\begin{figure*}[!htbp]
      \centering
      \includegraphics[width=0.85\linewidth]{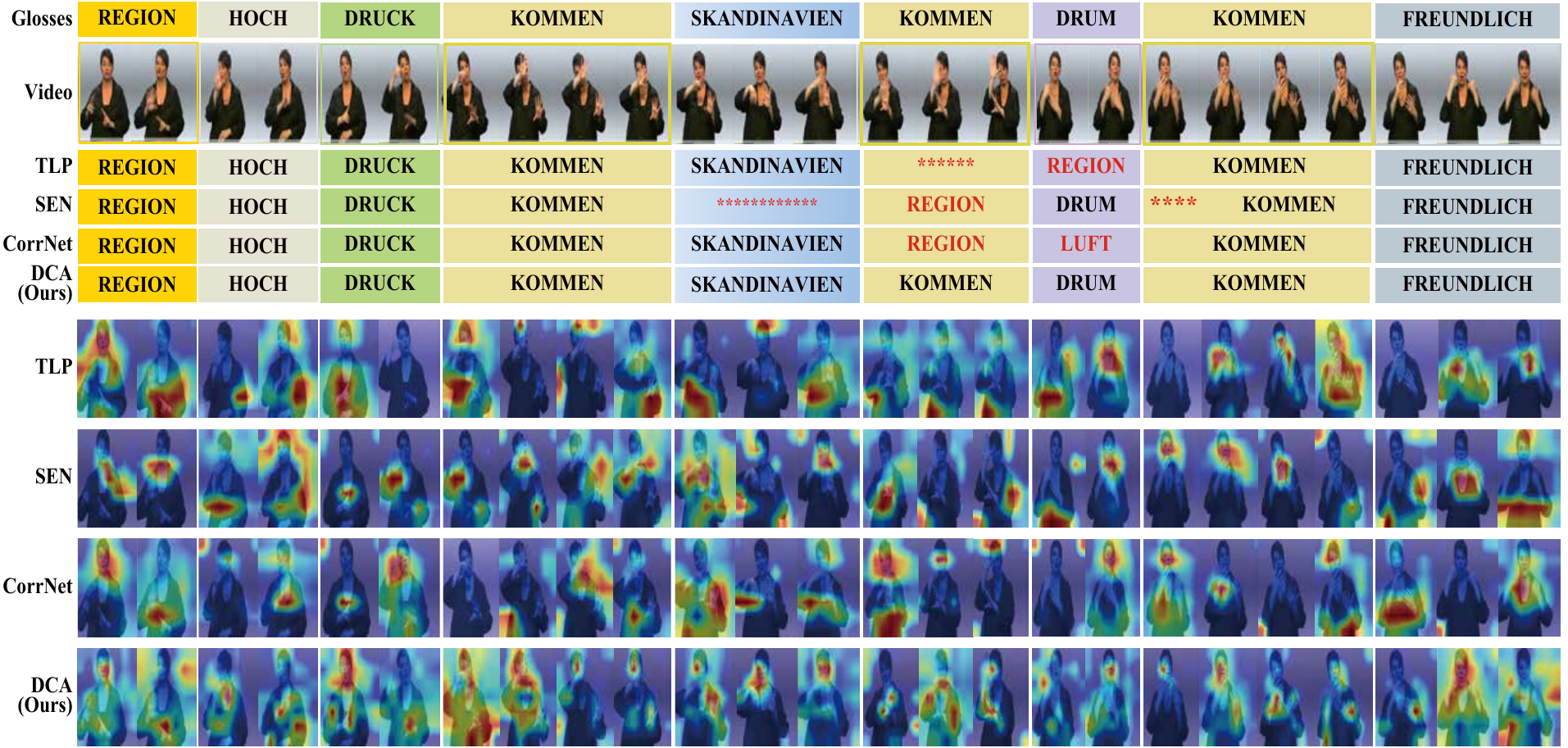}
      \caption{Visualizing the recognition results and the Grad-CAMs~\cite{gradcam2017} results of TLP~\cite{hu2022temporal}, SEN~\cite{Hu2022SelfEmphasizingNF}, CorrNet~\cite{hu2023continuous} and proposed DCA on a PHOENIX-2014 test video. 
      Glosses with red symbols denote the wrongly predicted gloss.
      The shades of color of the regions (blue, yellow, red, dark red) represent the weak to strong attention of the model to the sign spatial regions.}
      \label{fig:Gloss2}
\end{figure*}

\begin{figure}[!htbp]
      \centering
      \includegraphics[width=0.9\linewidth]{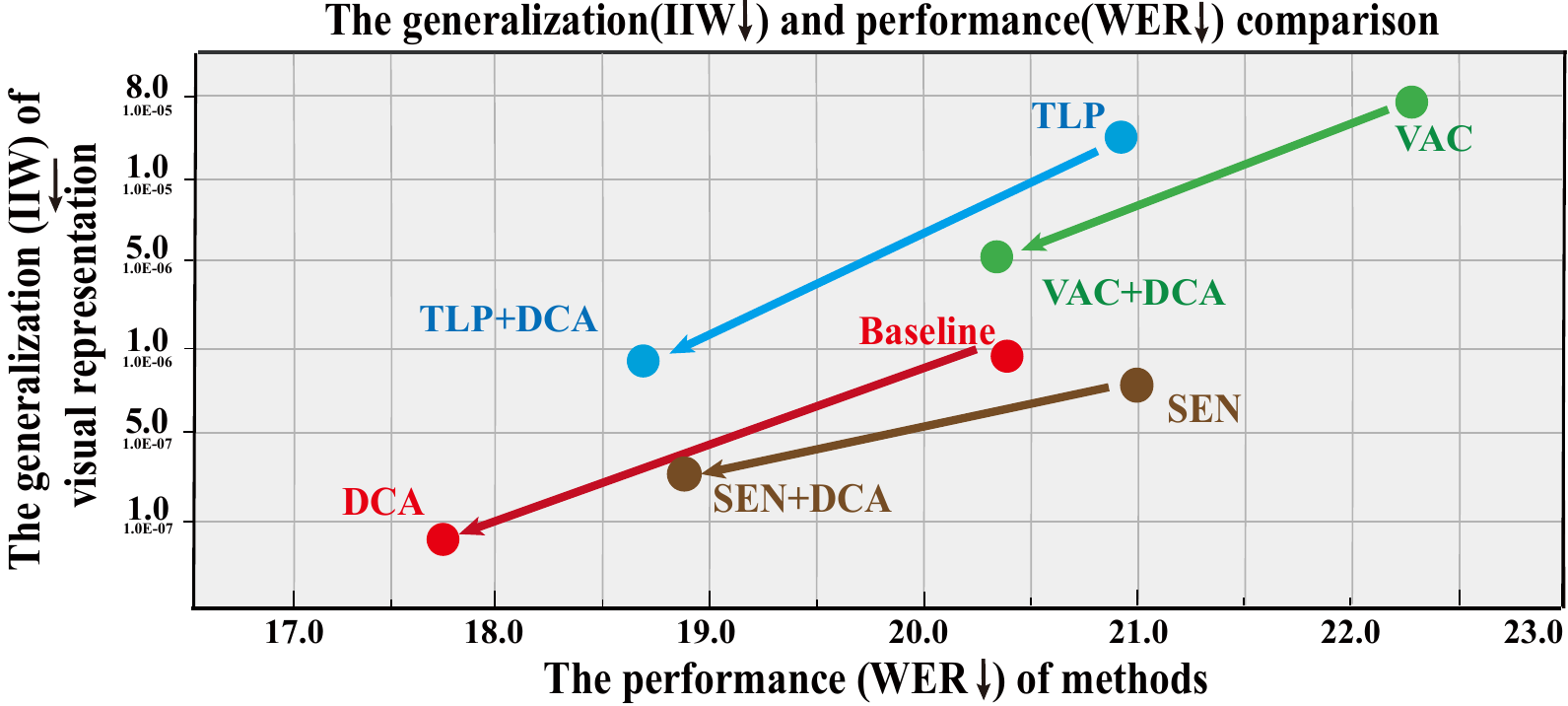}
      \caption{Evaluation for DCA's generalization and recognition accuracy over other SOTA CSLR methods on the PHOENIX-2014 test set.}
      \label{fig:DCA_gen}
\end{figure}

\subsection{Other Evaluations}
\noindent \textbf{Evaluation of global temporal context learning.}
In this section, we aim to evaluate the global temporal context learning of our DCA.
Specifically, we first extract the video sequence representation from two random videos of the Phoenix-2014 test set.
We then choose a window size to approximate the clip length of the video sequence in relation to the glosses. 
For example, if the video has 55 clips corresponding to 16 glosses, the window size is set to (55 mod 16 =3).
Next, we compute the cosine similarity between the cropped video clip representations. This allows each video clip to correspond roughly to a gloss (with “\textcolor{red}{***}” indicating unrecognizable glosses), and the cosine similarity results reflect the global temporal context correlation.
As shown in Figure~\ref{fig:DCAcontext}, our DCA effectively establishes higher similarity for similar glosses, with larger similarity differences between distinct glosses, and achieves almost the right corresponding with glosses.
Specifically, clips 4,5,7,9 correspond to the same gloss~(see red boxes), our DCA assigns them higher similarity than other methods.
Additionally, clip 3 has distinct gloss semantic compared to clip 1, and DCA shows a larger dissimilarity between these clips than the other methods. 
These results demonstrate that our DCA is capable of effectively learning the global temporal context.

\noindent \textbf{Evaluation on DCA generalizing other CSLR methods.}
We employ the compression of information
stored in weights~(IIW)~\cite{PIB2022} to quantitatively measure the generalization capability of our DCA.
As shown in~Figure~\ref{fig:DCA_gen}, we observe that with DCA optimization, all methods consistently experience an improvement in both performance and generalizability, with WER and IIW values decreasing significantly.
Specifically, the VAC~\cite{min2021visual} and TLP~\cite{hu2022temporal} gain remarkable enhancement in performance and generalizability.
In particular, TLP and SEN achieve impressive recognition results~(18.6\% and 18.8\% WERs) on the Phoenix-2014 test set.


\noindent \textbf{Method efficiency comparison.}
In this section, we employ the THOP~\cite{thop} tool to evaluate the parameters and GFLOPs of CSLR methods and evaluate their Throughout~(videos/s) in the inference process. 
We adopt the $140$ frames as the default.
As shown in Table~\ref{Table:effect}, our DCA achieves a balance between recognition performance and inference speed~(see Throughout).

\setlength{\tabcolsep}{1pt}
\begin{table}[h]
\centering
\fontsize{9}{12}\selectfont
\caption{Efficiency comparison between our DCA and other CSLR methods on the {PHOENIX-2014}. All experiments are measured on an A100 GPU with batch size 1.}
\begin{tabular}{c|ccc|cc} 
\toprule
        Methods & Param & GFLOPs & Throughout & Dev~(\%)~$\downarrow$ & Test~(\%)~$\downarrow$ \\ \midrule 
        VAC  &34.3& 567 & 17.0 & 21.2 & 22.3 \\  
         TLP &59.5& 573 & 17.0 & 19.7 & 20.2 \\ 
         SEN  & 34.5  &578 & 15.5 & 19.5 & 21.0  \\   
        C$^2$$ST_{SW+WE}$  &78.2 & 1368 & 4.4 & 17.6 & 18.3 \\
        DCA~(Ours)  &70.1& 1655 & 10.5 & \textbf{17.3} & \textbf{17.7} \\

        \bottomrule
    \end{tabular}
    \label{Table:effect}
\end{table}


\noindent \textbf{Evaluation of the capture of sign language movements.}
To further qualitatively evaluate the capability of our DCA to capture sign language movements. 
In Figure~\ref{fig:Gloss2}, we visualize the recognition results and the Class Activation Mapping~(CAM)~\cite{gradcam2017} of TLP~\cite{hu2022temporal}, SEN~\cite{Hu2022SelfEmphasizingNF},  CorrNet~\cite{hu2023continuous}, and our DCA using an example video from the PHOENIX-2014 test set.
In this example, the verb gloss ``KOMMEN'' appears three times within the sentence, each time conveying different meanings corresponding to ``came'', ``come from'', and ``will come''.
The interpretation of ``KOMMEN'' changes depending on the context, requiring a revised global temporal context informed by the gloss's global context for accurate understanding.
Our DCA successfully recognizes all ``kommen'', while TLP, SEN, and CorrNet fail to correctly interpret the second instance, as they do not incorporate textual grammar to revise their global temporal context. 
Regarding the Class Activation Mapping, we observe that our DCA is more sensitive and accurate in capturing the sign movements associated with specific areas of the video, demonstrating its superior recognition capability compared to the other methods.

\section{Conclusion}
This study investigates the effect of utilizing textual grammar to refine the video representation of continuous sign language recognition (CSLR).
We propose a novel approach called \textbf{D}enoising-\textbf{C}ontrastive \textbf{A}lignment (DCA), which facilitates both instance correspondence and global context alignment between video and gloss sequences to model textual grammar.
DCA approximates the instance correspondence between signs and glosses in a contrastive manner, providing gloss semantics for the video encoder to refine its representation for each sign.
Furthermore, DCA conducts a denoising-diffusion alignment to align the video and gloss sequence representations.
This process enables the model to learn the gloss global context by generative denoising the gloss sequence representation from noise based on the video.  
This alignment helps the video encoder refine its understanding of the global temporal context.
%
Experimental results on publicly available CSLR benchmarks demonstrate that DCA achieves excellent recognition performance. 
Other ablation study results not only validate the effectiveness of the proposed Denoising-Diffusion Alignment but also highlight the potential of the denoising-diffusion model for improving visual representation learning.

\noindent\textbf{Limitations.} 
Compared with the baseline method, about 8 minutes per epoch training time overhead will be incurred for DCA, which faces excessive computational overhead.
We may still need long-term research to achieve the denoising-diffusion model to work robustly in visual feature extraction and replace the alignment process in CSLR.


 \bibliographystyle{IEEEtran.bst} \bibliography{main}

\newpage

 




\begin{IEEEbiography}[{\includegraphics[width=1in,height=1.25in,clip,keepaspectratio]{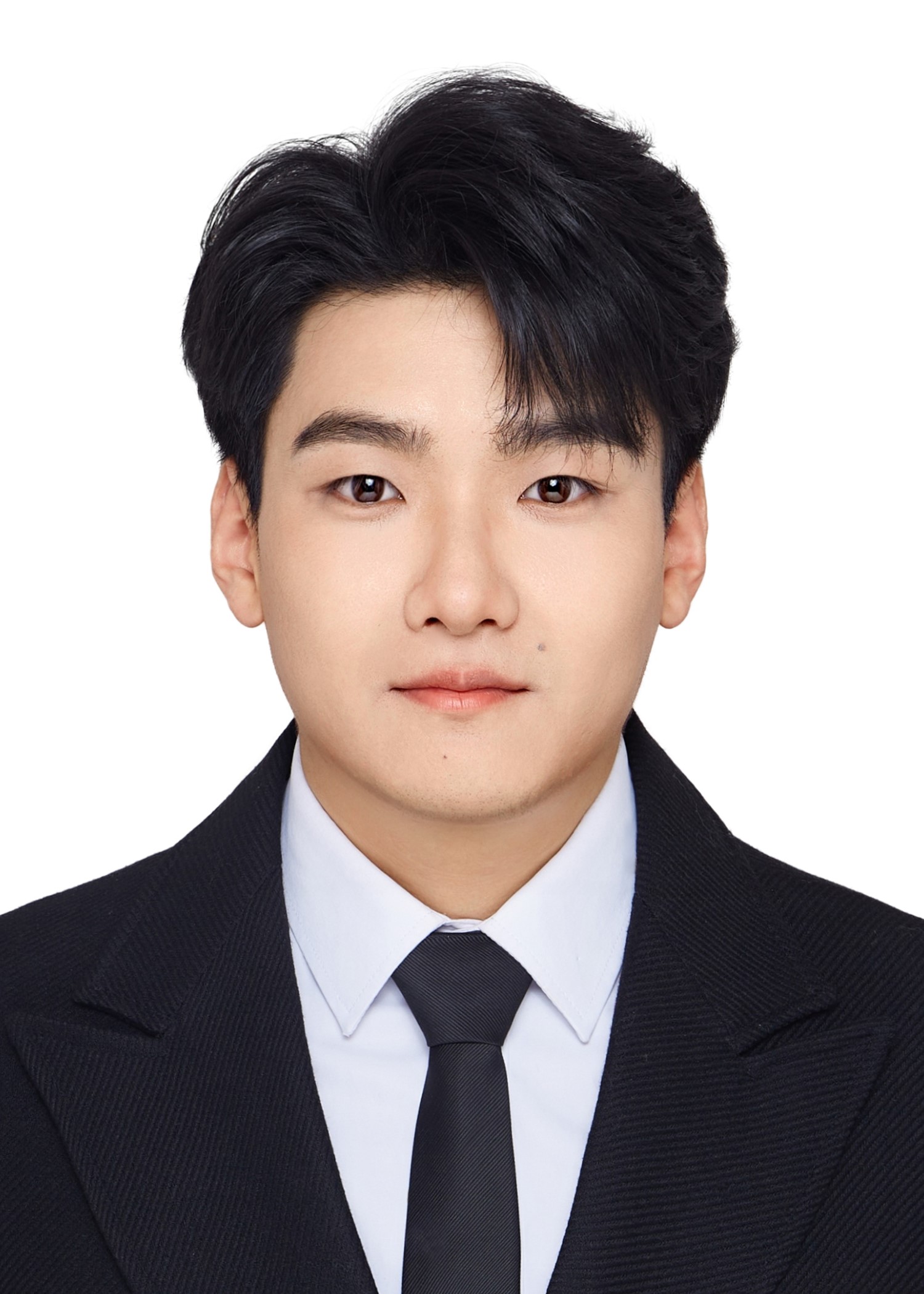}}]{Leming Guo} has received the Ph.D. degree in the computer science and technology from Tianjin University of Technology in 2024. He is currently a lecturer at the School of Computer Science and Engineering, Tianjin University of Technology. His research interests include sign language recognition, action recognition, domain adaptation, and multimedia analysis.
\end{IEEEbiography}
\vspace{-10 mm}
\begin{IEEEbiography}[{\includegraphics[width=1in,height=1.25in,clip,keepaspectratio]{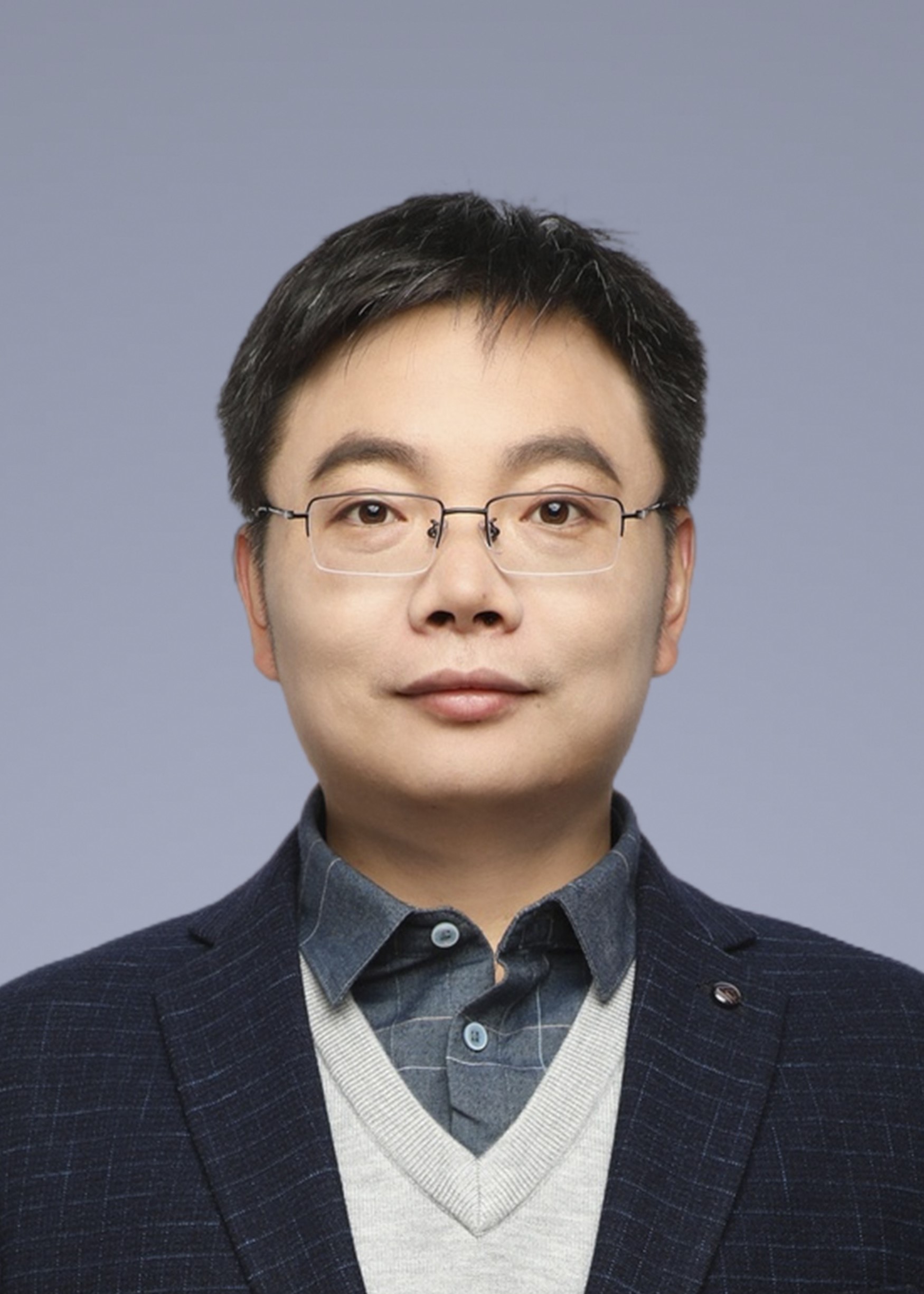}}]{Wanli Xue} is a professor of the school of computer science and engineering at Tianjin University of Technology. He received the Ph.D. degree in technology of computer application from Tianjin University in 2019. His research interests include visual tracking, images stitching, and sign language recognition.
\end{IEEEbiography}
\vspace{-10 mm}
\begin{IEEEbiography}[{\includegraphics[width=1in,height=1.25in,clip,keepaspectratio]{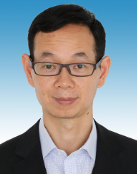}}]{Shengyong Chen} received the Ph.D. degree in robot vision from the City University of Hong Kong, Hong Kong, in 2003. From 2006 to 2007, he was with the University of Hamburg, Hamburg, Germany. He is currently a professor with the Tianjin University of Technology, Tianjin, China. His current research interests include computer vision, robotics, and image analysis.
\end{IEEEbiography}

\vfill

\end{document}